\documentclass{article} 
\usepackage{nips12submit_e,times}

\usepackage{bm}
\usepackage{graphicx}
\usepackage{amsfonts}

\newcommand{\be}{\begin{equation}}
\newcommand{\ee}{\end{equation}}

\newcommand{\bea}{\begin{eqnarray}}
\newcommand{\eea}{\end{eqnarray}}
\newcommand{\beastar}{\begin{eqnarray*}}
\newcommand{\eeastar}{\end{eqnarray*}}

\newcommand{\T}{^{\mathrm{T}}}
\newcommand{\tr}{\mathrm{tr}\,}

\newcommand{\Cf}{D}
\newcommand{\CF}{\bm{D}}
\newcommand{\Ps}{\bm{\Psi}}
\newcommand{\La}{\bm{\Lambda}}
\newcommand{\Ga}{\bm{\Gamma}}
\newcommand{\One}{\bm{I}}
\newcommand{\Gcal}{\mathcal{G}}
\newcommand{\ena}{\hat\epsilon}

\title{Learning curves for multi-task Gaussian process regression}
\author{Simon R F Ashton\\
King's College London\\
Department of Mathematics\\
Strand, London WC2R 2LS, U.K.\\
\And
Peter Sollich\\
King's College London\\
Department of Mathematics\\
Strand, London WC2R 2LS, U.K.\\
\texttt{peter.sollich@kcl.ac.uk}
}

\nipsfinalcopy

\begin{document}

\maketitle

\begin{abstract}
  We study the average case performance of multi-task Gaussian process (GP)
  regression as captured in the learning curve, i.e.\ the average Bayes error
  for a chosen task versus the total number of examples $n$ for all
  tasks. For GP covariances that are the product of an
  input-dependent covariance function and a free-form inter-task
  covariance matrix, we
  show that accurate approximations for the learning curve can be
  obtained
for an arbitrary number of tasks $T$.  We use
  these to study the asymptotic learning behaviour for large
  $n$. Surprisingly, multi-task learning can be asymptotically essentially
  useless, in the sense that examples from other tasks help only when the
  degree of inter-task correlation, $\rho$, is near its maximal value
  $\rho=1$. This effect is most extreme for learning of smooth target
  functions as described by e.g.\ squared exponential kernels. We also
  demonstrate that when learning {\em many} tasks, the learning curves
  separate into an initial phase, where the Bayes error on each task
  is reduced down to a plateau value by ``collective learning''
   even though most tasks have not seen examples,
  and a final decay that occurs once the number of examples is
  proportional to the number of tasks.
\end{abstract}

\section{Introduction and motivation}

Gaussian processes (GPs)~\cite{WilRas06} have been popular in the NIPS
community for a number of years now, as one of the key non-parametric
Bayesian inference approaches. In the simplest case one can use a GP
prior when learning a function from data. In line with growing
interest in multi-task or transfer learning, where relatedness between
tasks is used to aid learning of the individual tasks (see
e.g.~\cite{Baxter00,BenSch08}), GPs have increasingly also been used
in a multi-task setting. A number of different choices of
covariance functions have been proposed~\cite{TehSeeJor05,%
  BonAgaWil07,%
  BonChaWil08,%
  AlvLaw09,%
  LeePelKas11
}. These differ e.g.\ in assumptions on whether the functions to be
learned are related to a smaller number of latent functions or have
free-form inter-task correlations; for a recent review
see~\cite{AlvRosLaw12}.

Given this interest in multi-task GPs, one would like to quantify the
benefits that they bring compared to single-task learning.
PAC-style bounds for classification~\cite{Baxter00,BenSch08,Maurer06}
in more general multi-task scenarios exist, but there has been little
work on average case analysis. The basic question in this setting is:
how does the Bayes error on a given task depend on the number of
training examples for all tasks, when averaged over all data sets of
the given size. For a {\em single} regression task, this {\em learning
  curve} has become
relatively well understood since the late 1990s, with a number of bounds and approximations
available~\cite{OppViv99,TreWilOpp99,Sollich99,MalOpp01,%
MalOpp02,MalOpp02b,SolHal02,Sollich02c,%
Sollich05b
}
as well as some exact
predictions~\cite{UrrSol10}. Already {\em two-task} GP regression is
much more difficult to analyse, and
progress was made only very recently at NIPS 2009~\cite{Chai09}, where upper
and lower bounds for learning curves were derived. The tightest of
these bounds, however, either required evaluation by Monte Carlo
sampling, or assumed knowledge of the corresponding single-task
learning curves. Here our aim is to obtain accurate learning curve
approximations that apply to an {\em arbitrary} number $T$ of tasks, and
that can be evaluated explicitly without recourse to sampling.

We begin (Sec.~\ref{sec:eg_form}) by expressing the Bayes error for
any single task in a multi-task GP regression problem in a convenient
feature space form, where individual training examples enter
additively.  This requires the introduction of a non-trivial tensor structure
combining feature space components and tasks. Considering the change in error when adding an example
for some task leads to partial differential equations linking the
Bayes errors for all tasks. Solving these using the method of
characteristics then gives, as our primary result, the desired
learning curve approximation (Sec.~\ref{sec:solution}).  In
Sec.~\ref{sec:evaluation} we discuss some of its predictions. The
approximation correctly delineates the limits of pure transfer
learning, when all examples are from tasks other than the one of
interest. Next we compare with numerical simulations for some two-task
scenarios, finding good qualitative agreement. These results also
highlight a surprising feature, namely that asymptotically the
relatedness between tasks can become much less useful. We analyse this
effect in some detail, showing that it is most extreme for
learning of smooth functions. Finally we discuss the case of many
tasks, where there is an unexpected separation of the learning curves
into a fast initial error decay arising from ``collective learning'', and a
much slower final part where tasks are learned almost independently.

\section{GP regression and Bayes error}
\label{sec:eg_form}

We consider GP regression for $T$ functions $f_\tau(x)$,
$\tau=1,2,\ldots,T$. These functions have to be learned from $n$
training examples $(x_\ell, \tau_\ell, y_\ell)$, $\ell=1,\ldots,n$. Here $x_\ell$ is the training
input, $\tau_\ell\in\{1,\ldots,T\}$ denotes which task the example
relates to, and $y_\ell$ is the corresponding training output. We
assume that the latter is given by the target function value
$f_{\tau_\ell}(x_\ell)$ corrupted by i.i.d.\ additive Gaussian noise
with zero mean and variance $\sigma_{\tau_\ell}^2$. This setup allows the
noise level $\sigma_\tau^2$ to depend on the task.

In GP regression the prior over the functions $f_\tau(x)$ is a
Gaussian process. This means that for any set of inputs $x_\ell$ and
task labels $\tau_\ell$, the function values
$\{f_{\tau_\ell}(x_\ell)\}$ have a joint Gaussian distribution. As is
common we assume this to have zero mean, so the multi-task GP is fully
specified by the covariances $\langle
f_\tau(x)f_{\tau'}(x')\rangle=\mathcal{C}(\tau,x,\tau',x')$.  For this
covariance we take the flexible form from~\cite{BonAgaWil07},
$\langle f_\tau(x)f_{\tau'}(x')\rangle = \Cf_{\tau\tau'}
C(x,x')$. Here $C(x,x')$ determines the covariance between function
values at different input points, encoding ``spatial'' behaviour such
as smoothness and the lengthscale(s) over which the functions vary,
while the matrix $\CF$ is a free-form inter-task covariance matrix.

One of the attractions of GPs for regression is that, even though they
are non-parametric models with (in general) an infinite number of degrees
of freedom, predictions can be made in closed form, see
e.g.~\cite{WilRas06}. For a test point $x$ for task $\tau$, one would
predict as output the mean of $f_\tau(x)$ over the (Gaussian) posterior, which
is $\bm{y}\T\bm{K}^{-1}\bm{k}_\tau(x)$. Here $\bm{K}$ is the $n\times n$ Gram
matrix with entries $K_{\ell m} = \Cf_{\tau_\ell\tau_m} C(x_\ell,x_m)
+ \sigma_{\tau_\ell}^2\delta_{\ell m}$, while $\bm{k}_\tau(x)$ is a
vector with the $n$ entries $k_{\tau,\ell} =
\Cf_{\tau_\ell\tau}C(x_\ell,x)$. The error bar would be taken as the
square root of the posterior variance of $f_\tau(x)$, which is
\be
V_\tau(x) = \Cf_{\tau\tau}C(x,x) - \bm{k}_\tau\T(x)\bm{K}^{-1} \bm{k}_\tau(x)
\label{V_tau}
\ee
The learning curve for
task $\tau$ is defined as the mean-squared prediction error,
averaged over the location of test input $x$ and over all
data sets with a specified number of examples for each task, say $n_1$
for task 1 and so on. As is standard in learning curve analysis we
consider a matched scenario where the training outputs $y_\ell$ are
generated from the same prior and noise model that we use for
inference. In this case the mean-squared prediction error
$\ena_\tau$ is the Bayes error, and is given by the average
posterior variance~\cite{WilRas06}, i.e.\ $\ena_\tau =
\langle V_\tau(x)\rangle_x$. To obtain the learning curve this is 
averaged over the location of the
training inputs $x_\ell$: $\epsilon_\tau = \langle
\ena_\tau\rangle$. This average presents the main challenge for learning
curve prediction because the training inputs feature in a highly nonlinear
way in $V_\tau(x)$.  Note that the training outputs, on the other hand, do
not appear in the posterior variance $V_\tau(x)$ and so do not
need to be averaged over.

We now want to write the Bayes error $\ena_\tau$ in a form convenient
for performing, at least approximately, the averages required for the
learning curve. Assume that all training inputs $x_\ell$, and also the
test input $x$, are drawn from the same distribution $P(x)$. One can
decompose the input-dependent part of the covariance function into
eigenfunctions relative to $P(x)$, according to $C(x,x')=\sum_i
\lambda_i \phi_i(x)\phi_i(x')$. The eigenfunctions are defined by the
condition $\langle C(x,x')\phi_i(x')\rangle_{x'} = \lambda_i
\phi_i(x)$ and can be chosen to be orthonormal with respect to $P(x)$,
$\langle \phi_i(x)\phi_j(x)\rangle_x = \delta_{ij}$. The sum over $i$
here is in general infinite (unless the covariance function is
degenerate, as e.g.\ for the dot product kernel $C(x,x')=x\cdot
x'$). To make the algebra below as simple as possible, we let the
eigenvalues $\lambda_i$ be arranged in decreasing order and truncate
the sum to the finite range $i=1,\ldots,M$; $M$ is then some large effective
feature space dimension and can be taken to infinity at the end.

In terms of the above eigenfunction decomposition, the Gram matrix has elements 
\[
K_{\ell m} = \Cf_{\tau_\ell\tau_m}
\sum_i \lambda_i \phi_i(x_\ell)\phi_i(x_m) + \sigma_{\tau_\ell}^2\delta_{\ell m}
=\sum_{i,\tau,j,\tau'} \delta_{\tau_\ell,\tau} \phi_i(x_\ell)
\lambda_i \delta_{ij} \Cf_{\tau\tau'} \phi_j(x_m)
\delta_{\tau',\tau_m}
 + \sigma_{\tau_\ell}^2\delta_{\ell m}
\]
or in matrix form $\bm{K} = \Ps \bm{L} \Ps\T + \bm{\Sigma}$ where $\bm{\Sigma}$
is the diagonal matrix from the noise variances and
\be
\Psi_{\ell,i\tau} = \delta_{\tau_\ell,\tau} \phi_i(x_\ell), \qquad
L_{i\tau,j\tau'} = \lambda_i \delta_{ij} \Cf_{\tau\tau'}
\label{Psi_L_def}
\ee
Here $\Ps$ has its second index ranging over $M$ (number of kernel
eigenvalues) times $T$ (number of tasks) values; $\bm{L}$ is a square
matrix of this size. In Kronecker (tensor) product notation,
$\bm{L}=\CF\otimes \La$ if we define $\La$ as the diagonal matrix with entries
$\lambda_i \delta_{ij}$. The Kronecker product is convenient for the
simplifications below; we will use that for generic square matrices,
$(\bm{A}\otimes \bm{B})(\bm{A'}\otimes \bm{B'}) =
(\bm{A}\bm{A'})\otimes(\bm{B}\bm{B'})$, $(\bm{A}\otimes \bm{B})^{-1} =
\bm{A}^{-1}\otimes \bm{B}^{-1}$, and $\tr(\bm{A}\otimes \bm{B}) =
(\tr\bm{A})(\tr\bm{B})$. In thinking about the mathematical
expressions, it is often easier to picture Kronecker products over
feature spaces and tasks as block matrices. For example, $\bm{L}$ can
then be viewed as consisting of $T\times T$ blocks, each of which is
proportional to $\La$.

To calculate the Bayes error, we need to average the posterior
variance $V_\tau(x)$ over the test input $x$.
The first term in (\ref{V_tau}) then becomes
$\Cf_{\tau\tau} \langle C(x,x)\rangle = \Cf_{\tau\tau}\tr\La$.
In the second one, we need to average
\beastar
\langle k_{\tau,\ell}(x) k_{\tau,m}\rangle_x &=&
\Cf_{\tau\tau_\ell} \langle C(x_\ell,x)C(x,x_m)\rangle_x
\Cf_{\tau_m\tau} \\
&=& \Cf_{\tau\tau_\ell} \sum_{ij}\lambda_i \lambda_j \phi_i(x_\ell)\langle
\phi_i(x) \phi_j(x)\rangle_x \phi_j(x_m) \Cf_{\tau_m\tau}
\\&=&
\sum_{i,\tau',j,\tau''} \Cf_{\tau \tau'} \Psi_{l,i\tau'} \lambda_i
\lambda_j \delta_{ij} \Psi_{m,j\tau''} \Cf_{\tau''\tau}
\eeastar
In matrix form this is
$
\langle \bm{k}_{\tau}(x) \bm{k}_{\tau}\T(x)\rangle_x = \Ps
[(\CF \bm{e}_\tau \bm{e}_\tau\T \CF)\otimes \La^2] \Ps\T = \Ps \bm{M}_\tau\Ps\T
$
Here the last equality defines $\bm{M}_\tau$, and we have denoted by
$\bm{e}_\tau$ the $T$-dimensional vector with $\tau$-th component
equal to one and all others zero. Multiplying by the inverse Gram matrix
$\bm{K}^{-1}$ and taking the trace gives the average of the second
term in (\ref{V_tau}); combining with the first gives
the Bayes error on task $\tau$
\[
\ena_\tau = \langle V_\tau(x)\rangle_x = \Cf_{\tau\tau} \tr\La
- \tr \Ps \bm{M}_\tau\Ps\T(\Ps \bm{L} \Ps\T + \bm{\Sigma})^{-1}
\]
Applying the Woodbury identity and re-arranging yields
\beastar
\ena_\tau
&=& \Cf_{\tau\tau} \tr\La
- \tr \bm{M}_\tau\Ps\T\bm{\Sigma}^{-1}\Ps (\One + \bm{L}\Ps\T\bm{\Sigma}^{-1}\Ps)^{-1}
\\
&=& \Cf_{\tau\tau} \tr \La
- \tr \bm{M}_\tau\bm{L}^{-1} [\One - (\One + \bm{L}\Ps\T\bm{\Sigma}^{-1}\Ps)^{-1}]
\eeastar
But
\beastar
\tr \bm{M}_\tau\bm{L}^{-1} &=&
\tr\{[(\CF \bm{e}_\tau \bm{e}_\tau\T \CF)\otimes \La^2]
[\CF\otimes \La]^{-1}\} \\
&=& \tr\{[\CF \bm{e}_\tau \bm{e}_\tau\T]\otimes \La\}
\ = \ \bm{e}_\tau\T \CF \bm{e}_\tau \,\tr\La = \Cf_{\tau\tau}\,\tr\La
\eeastar
so the first and second terms in the expression for $\ena_\tau$
cancel and one has
\beastar
\ena_\tau
&=& \tr \bm{M}_\tau\bm{L}^{-1}(\One + \bm{L}\Ps\T\bm{\Sigma}^{-1}\Ps)^{-1}
\ = \ \tr \bm{L}^{-1} \bm{M}_\tau\bm{L}^{-1}(\bm{L}^{-1} +
\Ps\T\bm{\Sigma}^{-1}\Ps)^{-1}
\\
&=& \tr [\CF\otimes \La]^{-1}[(\CF \bm{e}_\tau \bm{e}_\tau\T \CF)\otimes \La^2]
[\CF\otimes \La]^{-1}
(\bm{L}^{-1} + \Ps\T\bm{\Sigma}^{-1}\Ps)^{-1}
\\
&=& \tr [\bm{e}_\tau \bm{e}_\tau\T \otimes \One]
(\bm{L}^{-1} + \Ps\T\bm{\Sigma}^{-1}\Ps)^{-1}
\eeastar
The matrix in square brackets in the last line is just a projector
$\bm{P}_\tau$ onto task $\tau$; thought of as a matrix of $T\times T$
blocks (each of size $M\times M$), this has an identity matrix in the
$(\tau,\tau)$ block while all other blocks are zero.  We can therefore
write, finally, for the Bayes error on task $\tau$,
\be
\ena_\tau
= \tr \bm{P}_\tau (\bm{L}^{-1} + \Ps\T\bm{\Sigma}^{-1}\Ps)^{-1}
\ee
Because $\bm{\Sigma}$ is diagonal and given the definition (\ref{Psi_L_def}) of
$\bm{\Psi}$, the matrix $\Ps\T\bm{\Sigma}^{-1}\Ps$ is a
{\em sum} of contributions from the individual training examples
$\ell=1,\ldots,n$. This will be important for deriving the learning
curve approximation below.
We note in passing that, because $\sum_\tau \bm{P}_\tau=\One$, the
sum of the Bayes errors on all tasks is 
$\sum_\tau \ena_\tau = \tr (\bm{L}^{-1} +
\Ps\T\bm{\Sigma}^{-1}\Ps)^{-1}$, in close analogy to the
corresponding expression for the single-task case~\cite{Sollich99}.

\section{Learning curve prediction}
\label{sec:solution}

To obtain the learning curve $\epsilon_\tau = \langle
\ena_\tau\rangle$, we now need to carry out the average $\langle
\ldots \rangle$ over the
training inputs. To help with this, we can extend an approach for the
single-task scenario~\cite{Sollich99} and define a response or
resolvent matrix
%
$
\Gcal = (\bm{L}^{-1} + \Ps\T\bm{\Sigma}^{-1}\Ps + \sum_\tau v_\tau
\bm{P}_\tau )^{-1}
$
%
with auxiliary parameters $v_\tau$ that will be set back to zero at
the end. One can then ask how $\bm{G}=\langle \Gcal\rangle$ and hence
$\epsilon_{\tau'}=\tr \bm{P}_{\tau'}\bm{G}$ changes
with the number $n_\tau$ of training points for task $\tau$. Adding an
example at position $x$ for task $\tau$ increases
$\Ps\T\bm{\Sigma}^{-1}\Ps$ by $\sigma_{\tau}^{-2} \bm{\phi}_\tau
\bm{\phi}_\tau\T$, where $\bm{\phi}_\tau$ has elements
$(\phi_\tau)_{i\tau'} = \phi_i(x)\delta_{\tau\tau'}$. Evaluating the
difference $(\Gcal^{-1}+\sigma_{\tau}^{-2} \bm{\phi}_\tau
\bm{\phi}_\tau\T)^{-1}-\Gcal$ with the help of the Woodbury identity
and approximating it with a derivative gives
\[
\frac{\partial\Gcal}{\partial n_\tau} = -
\frac{\Gcal\bm{\phi}_\tau\bm{\phi}_\tau\T \Gcal}
{\sigma_\tau^{2}+\bm{\phi}_\tau\T \Gcal \bm{\phi}_\tau}
\]
This needs to be averaged over the new example and all previous
ones. If we approximate by averaging numerator and denominator
separately we get
\be
\frac{\partial \bm{G}}{\partial n_\tau} = 
\frac{1}
{\sigma_\tau^{2}+\tr\bm{P}_\tau \bm{G}}
\frac{\partial \bm{G}}{\partial v_\tau}
\label{Gmatrix_eq}
\ee
Here we have exploited for the average over $x$ that the matrix
$\langle \bm{\phi}_\tau\bm{\phi}_\tau\T\rangle_x$ has
$(i,\tau'),(j,\tau'')$-entry $\langle \phi_i(x)\phi_j(x)\rangle_x
\delta_{\tau\tau'} \delta_{\tau\tau''}=\delta_{ij}\delta_{\tau\tau'}
\delta_{\tau\tau''}$, hence simply $\langle
\bm{\phi}_\tau\bm{\phi}_\tau\T\rangle_x = \bm{P}_\tau$. 
We have also used the auxiliary parameters to rewrite
$-\langle \Gcal \bm{P}_{\tau} \Gcal\rangle =
\partial\langle \Gcal\rangle/\partial v_\tau = 
\partial\bm{G}/\partial v_\tau$.
Finally,
multiplying (\ref{Gmatrix_eq}) by $\bm{P}_{\tau'}$ and taking the
trace gives the set of quasi-linear partial differential equations
\be
\frac{\partial \epsilon_{\tau'}}{\partial n_\tau} = 
\frac{1}
{\sigma_\tau^{2}+\epsilon_\tau}
\frac{\partial \epsilon_{\tau'}}{\partial v_\tau}
\label{eps_PDEs}
\ee
The remaining task is now to find the functions
$\epsilon_\tau(n_1,\ldots,n_T,v_1,\ldots,v_T)$ by solving these
differential equations. We initially attempted to do this by tracking
the $\epsilon_\tau$ as examples are added one task at a time, but the
derivation is laborious already for $T=2$ and becomes prohibitive beyond.
Far more elegant is to adapt the method of characteristics to
the present case. We need to find a $2T$-dimensional surface in the
$3T$-dimensional space
$(n_1,\ldots,n_T,v_1,\ldots,v_T,\epsilon_1,\ldots,\epsilon_T)$, which
is specified by the $T$ functions $\epsilon_\tau(\ldots)$.  A small
change $(\delta n_1,\ldots,\delta n_T,\delta v_1,\ldots,\delta
v_T,\delta \epsilon_1,\ldots,\delta \epsilon_T)$ in all $3T$
coordinates is tangential to this surface if it obeys the $T$
constraints (one for each $\tau'$)
\[
\delta\epsilon_{\tau'} = \sum_\tau \left(
\frac{\partial \epsilon_{\tau'}}{\partial n_\tau} \delta n_\tau + 
\frac{\partial \epsilon_{\tau'}}{\partial v_\tau} \delta v_\tau
\right)
\]
From (\ref{eps_PDEs}), one sees that this condition is satisfied
whenever
$
\delta\epsilon_\tau = 0$ 
and $\delta n_\tau = - \delta v_\tau
(\sigma_\tau^2 + \epsilon_\tau)
$
It follows that all the characteristic curves given by
$\epsilon_\tau(t) = \epsilon_{\tau,0}={\rm const.}$,
$v_\tau(t) = v_{\tau,0}(1-t)$, $n_\tau(t) =
v_{\tau,0}(\sigma_\tau^2+\epsilon_{\tau,0})\,t$ 
for $t\in[0,1]$ are tangential to the solution surface for all $t$, so
lie within this surface if the initial point at $t=0$ does. Because at
$t=0$ there are no training examples ($n_\tau(0)=0$), this initial
condition is satisfied by setting
\[
\epsilon_{\tau,0} = \tr \bm{P}_\tau \left(\bm{L}^{-1} + \sum_{\tau'}
  v_{\tau',0} 
\bm{P}_{\tau'} \right)^{-1}
\]
Because $\epsilon_\tau(t)$ is constant along the characteristic curve,
we get by equating the values at $t=0$ and $t=1$
\[
\epsilon_{\tau,0}= 
\tr \bm{P}_\tau \left(
\bm{L}^{-1} + \sum_{\tau'} v_{\tau',0} \bm{P}_{\tau'}
\right)^{-1}=
\epsilon_\tau(\{n_{\tau'}=v_{\tau',0}(\sigma_{\tau'}^2+\epsilon_{\tau',0})\},
\{v_{\tau'}=0\})
\]
Expressing $v_{\tau',0}$ in terms of $n_{\tau'}$ gives then
\be
\epsilon_{\tau}= 
\tr \bm{P}_\tau \left(\bm{L}^{-1} + \sum_{\tau'}
\frac{n_{\tau'}}{\sigma_{\tau'}^2+\epsilon_{\tau'}} \bm{P}_{\tau'}\right)^{-1}
\label{eps_sol}
\ee
This is our main result: a closed set of $T$ self-consistency
equations for the average Bayes errors $\epsilon_\tau$. Given $\bm{L}$ as
defined by the eigenvalues $\lambda_i$ of the covariance function, the
noise levels $\sigma_\tau^2$ and the number of examples $n_\tau$ for
each task, it is straightforward to solve these equations numerically
to find the average Bayes error $\epsilon_\tau$ for each task. 

The r.h.s.\ of (\ref{eps_sol}) is easiest to evaluate if we view 
the matrix inside the brackets as consisting of $M\times M$ blocks of
size $T\times T$ (which is the reverse of the picture we have used so
far). The matrix is then block diagonal, with the blocks
corresponding to different eigenvalues $\lambda_i$. Explicitly,
because $\bm{L}^{-1} = \CF^{-1}\otimes \La^{-1}$, one has
\be
\epsilon_{\tau}= 
\sum_i \biggl(\lambda_i^{-1}\CF^{-1} + 
{\rm diag}(\{\frac{n_{\tau'}}{\sigma_{\tau'}^2+\epsilon_{\tau'}}\})\biggr)^{-1}_{\tau\tau}
\label{eps_sol2}
\ee

\section{Results and discussion}
\label{sec:evaluation}

We now consider the consequences of the approximate prediction
(\ref{eps_sol2}) for multi-task learning curves in GP regression. A
trivial special case is the one of uncorrelated tasks, where
$\bm{D}$ is diagonal. Here one recovers $T$ separate equations for the
individual tasks as expected, which have the same form as 
for single-task learning~\cite{Sollich99}.

\subsection{Pure transfer learning}

Consider now the case of pure transfer learning, where one is learning
a task of interest (say $\tau=1$) purely from examples for other
tasks. What is the lowest average Bayes error that can be obtained?
Somewhat more generally, suppose we have no examples for the first
$T_0$ tasks, $n_1=\ldots =n_{T_0}=0$, but a large number of examples for
the remaining $T_1=T-T_0$ tasks. Denote $\bm{E}=\CF^{-1}$ and write this
in block form as
\[
\bm{E} = \left( \begin{array}{cc} \bm{E}_{00} & \bm{E}_{01} \\
    \bm{E}_{01}\T & \bm{E}_{11} \end{array}\right)
\]
Now multiply by $\lambda_i^{-1}$ and add in the lower right block a
diagonal matrix
$\bm{N}=\mbox{diag}(\{n_{\tau}/(\sigma_{\tau}^2+\epsilon_\tau)\}_{\tau=T_0+1,\ldots,T})$.
The matrix inverse in (\ref{eps_sol2}) then has top left block
$\lambda_i[\bm{E}_{00}^{-1} + \bm{E}_{00}^{-1}\bm{E}_{01}(\lambda_i
\bm{N} + \bm{E}_{11} -
\bm{E}_{01}\T\bm{E}_{00}^{-1}\bm{E}_{01})^{-1}\bm{E}_{01}\T\bm{E}_{00}^{-1}]$. As
the number of examples for the last $T_1$ tasks grows, so do all
(diagonal) elements of $\bm{N}$. In the limit only the term
$\lambda_i\bm{E}_{00}^{-1}$ survives, and summing over $i$ gives
$\epsilon_1 = \tr\La(\bm{E}_{00}^{-1})_{11} = \langle C(x,x)\rangle
(\bm{E}_{00}^{-1})_{11}$. The Bayes error on task 1 cannot become
lower than this, placing a limit on the benefits of pure transfer
learning. That this prediction of the approximation (\ref{eps_sol2})
for such a lower limit is correct can also be checked directly: once
the last $T_1$ tasks $f_\tau(x)$ ($\tau=T_0+1,\ldots T$) have been
learn perfectly, the posterior over the first $T_0$ functions is, by
standard Gaussian conditioning, a GP with covariance
$C(x,x')(\bm{E}_{00})^{-1}$. Averaging the posterior variance of
$f_1(x)$ then gives the Bayes error on task 1 as $\epsilon_1 = \langle
C(x,x)\rangle (\bm{E}_{00}^{-1})_{11}$, as found earlier.

This analysis can be extended to the case where there are some examples
available also for the first $T_0$ tasks.
One finds for the generalization errors on these tasks the prediction
(\ref{eps_sol2}) with $\bm{D}^{-1}$ replaced by
$\bm{E}_{00}$. This is again in line with the above form of the GP posterior
after perfect
learning of the remaining $T_1$ tasks.
%

\subsection{Two tasks}

We next analyse how well the approxiation (\ref{eps_sol2}) does in
predicting multi-task learning curves for $T=2$ tasks. Here we have
the work of Chai~\cite{Chai09} as a baseline, and as there we choose 
\[
\CF = \left( \begin{array}{cc} 1 & \rho \\ \rho & 1 \end{array}\right)
\]
The diagonal elements are fixed to unity, as in a practical
application where one would scale both task functions $f_1(x)$ and
$f_2(x)$ to unit variance; the degree of correlation of the tasks
is controlled by $\rho$. We fix $\pi_2=n_2/n$ and plot learning curves
against $n$. In numerical simulations we ensure integer values of
$n_1$ and $n_2$ by setting $n_2=\lfloor n\pi_2\rfloor$, $n_1=n-n_2$;
for evaluation of (\ref{eps_sol2}) we use directly $n_2=n\pi_2$,
$n_1=n(1-\pi_2)$. For simplicity we consider equal noise levels
$\sigma_1^2=\sigma_2^2=\sigma^2$.

\begin{figure}
\centering{\includegraphics[width=0.90\textwidth,clip]{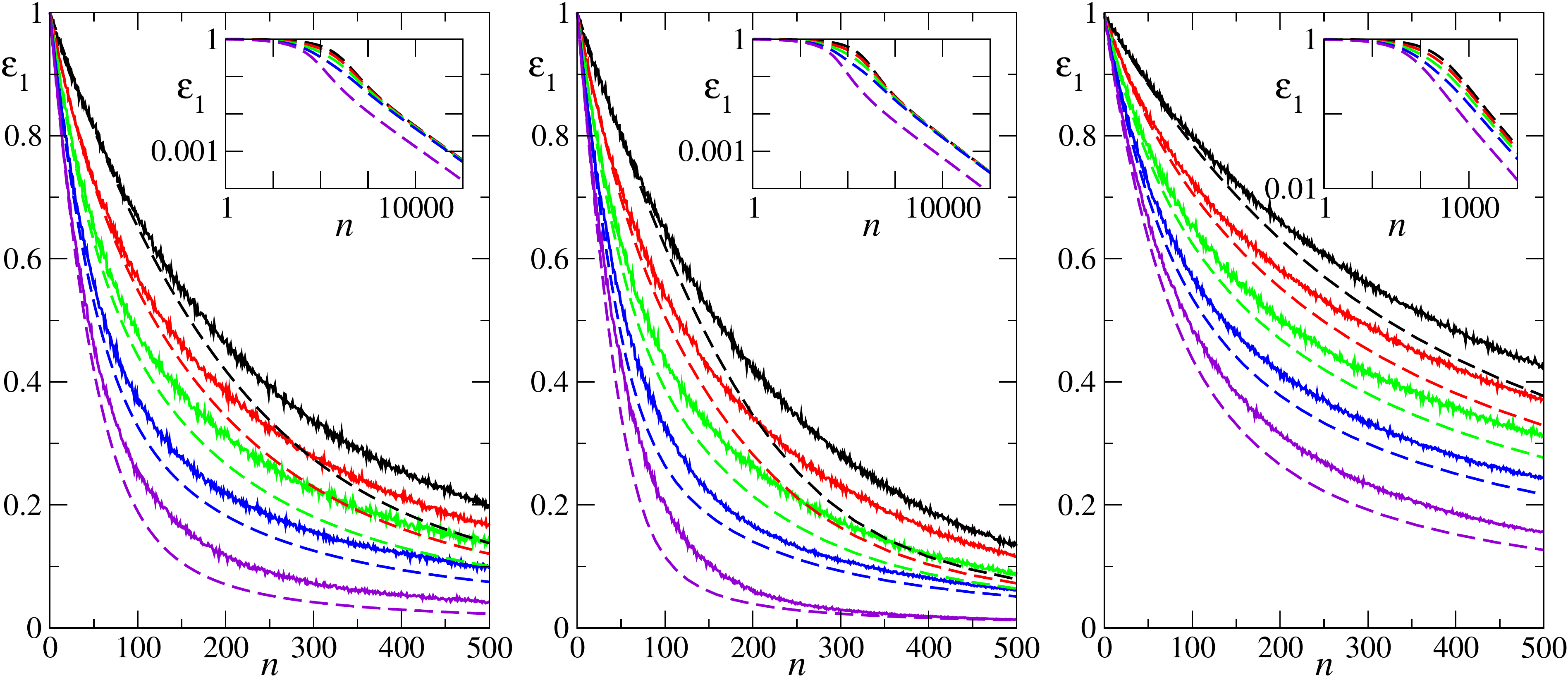}}
\caption{Average Bayes error for task 1 for two-task GP regression
  with kernel lengthscale $l=0.01$, noise level
  $\sigma^2=0.05$ and a fraction $\pi_2=0.75$ of examples for task
  2. Solid lines: numerical simulations; dashed lines: approximation
  (\ref{eps_sol2}). Task correlation $\rho^2=0$, 0.25, 0.5, 0.75, 1 from
  top to bottom. Left: SE covariance function, Gaussian input
  distribution. Middle: SE covariance, uniform inputs. Right: OU
  covariance, uniform inputs. Log-log plots (insets) show tendency of asymptotic uselessness,
  i.e.\ bunching of the $\rho<1$ curves towards the one for $\rho=0$;
  this effect is strongest for learning of smooth functions (left and middle).
\label{fig:SE_Gaussian}
\label{fig:SE_uniform}
\label{fig:OU_uniform}
}
\end{figure}

As regards the covariance function and input distribution, we analyse
first the scenario studied in~\cite{Chai09}: a squared
exponential (SE) kernel $C(x,x')=\exp[-(x-x')^2/(2l^2)]$ with
lengthscale $l$, and one-dimensional inputs $x$ with a Gaussian
distribution $\mathcal{N}(0,1/12)$. The kernel eigenvalues $\lambda_i$
are known explicitly from~\cite{ZhuWilRohMor98} and decay
exponentially with $i$. Figure~\ref{fig:SE_Gaussian}(left) compares
numerically simulated learning curves with the predictions for
$\epsilon_1$, the average Bayes error on task 1, from
(\ref{eps_sol2}). Five pairs of curves are shown, for $\rho^2=0$,
0.25, 0.5, 0.75, 1. Note that the two extreme values represent
single-task limits, where examples from task 2 are either ignored
($\rho=0$) or effectively treated as being from task 1 ($\rho=1$). Our
predictions lie generally below the true learning curves, but
qualitatively represent the trends well, in particular the variation
with $\rho^2$. The curves for the different $\rho^2$ values are fairly
evenly spaced vertically for small number of examples, $n$,
corresponding to a linear dependence on $\rho^2$. As $n$ increases,
however, the learning curves for $\rho<1$ start to bunch together and
separate from the one for the fully correlated case ($\rho=1$). The
approximation (\ref{eps_sol2}) correctly captures this behaviour,
which is discussed in more detail below. 

Figure~\ref{fig:SE_uniform}(middle) has analogous results for the case
of inputs $x$ uniformly distributed on the interval $[0,1]$; the
$\lambda_i$ here decay exponentially with
$i^2$~\cite{SolHal02}. Quantitative agreement between simulations and
predictions is better for this case. The discussion in~\cite{SolHal02} suggests
that this is because the approximation method we have used implicitly neglects
spatial variation of the dataset-averaged posterior variance $\langle
V_\tau(x)\rangle$; but for a uniform input distribution this variation
will be weak except near the ends of the input range
$[0,1]$. Figure~\ref{fig:OU_uniform}(right) displays similar results for an OU
kernel $C(x,x')=\exp(-|x-x'|/l)$, showing that our predictions also work
well when learning rough (nowhere differentiable) functions.

\subsection{Asymptotic uselessness}

The two-task results above suggest that multi-task learning is less
useful asymptotically: when the number of training examples $n$ is
large, the learning curves seem to bunch towards the curve for
$\rho=0$, where task 2 examples are ignored, except when the two tasks
are fully correlated ($\rho=1$). We now study this effect.

\begin{figure}
\centering{\includegraphics[width=0.9\textwidth,clip]{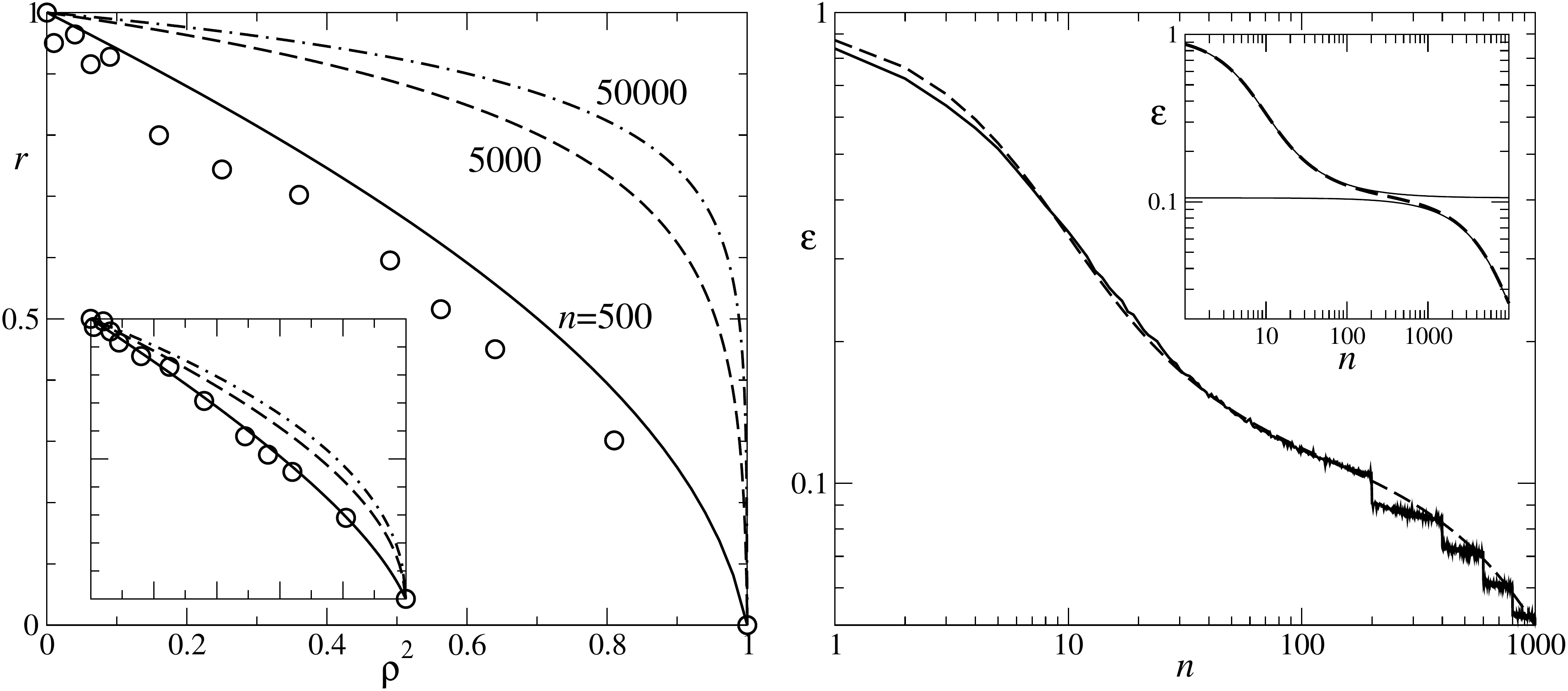}}
\caption{Left: Bayes error (parameters as in
  Fig.~\ref{fig:SE_Gaussian}(left), with $n=500$) vs $\rho^2$. To focus on the
  error reduction with $\rho$,
  $r=[\epsilon_1(\rho)-\epsilon_1(1)]/[\epsilon_1(0)-\epsilon_1(1)]$ is
  shown. Circles: simulations; solid line: predictions from
  (\ref{eps_sol2}). Other lines: predictions for
  larger $n$, showing the approach to asymptotic uselessness in
  multi-task learning of smooth functions. Inset: Analogous results
  for rough functions (parameters as in
  Fig.~\ref{fig:OU_uniform}(right)).
Right: Learning curve for many-task learning ($T=200$, parameters
otherwise as in Fig.~\ref{fig:SE_Gaussian}(left) except $\rho^2=0.8$). Notice the bend
around $\epsilon_1=1-\rho=0.106$. Solid line: simulations (steps arise
because we chose to allocate examples to tasks in order
$\tau=1,\ldots,T$ rather than randomly); dashed line: predictions from
(\ref{eps_sol2}). Inset: Predictions for
$T=1000$, with asymptotic forms $\epsilon=1-\rho+\rho\tilde\epsilon$ and
$\epsilon=(1-\rho)\bar\epsilon$ for the two learning stages 
shown as solid lines.
\label{fig:SE_useless}
\label{fig:OU_useless}
\label{fig:many_tasks}
}
\end{figure}

When the number of examples for all tasks becomes large, the Bayes
errors $\epsilon_\tau$ will become small and eventually be negligible
compared to the noise variances $\sigma_\tau^2$ in
(\ref{eps_sol2}). One then has an explicit prediction for each
$\epsilon_\tau$, without solving $T$ self-consistency equations. If we
write, for $T$ tasks, $n_\tau=n\pi_\tau$ with $\pi_\tau$ the fraction
of examples for task $\tau$, and set $\gamma_\tau =
\pi_\tau/\sigma_\tau^2$, then for large $n$
\be
\epsilon_{\tau}= 
{\textstyle\sum_i} \left(\lambda_i^{-1}\CF^{-1} + 
n \Ga\right)^{-1}_{\tau\tau} =
{\textstyle\sum_i} (\Ga^{-1/2}[\lambda_i^{-1}(\Ga^{1/2}\CF\Ga^{1/2})^{-1} +
  n\One]^{-1} \Ga^{-1/2})_{\tau\tau}
\label{eps_asympt1}
\ee
where $\Ga = \mbox{diag}(\gamma_1,\ldots,\gamma_T)$. Using an
eigendecomposition of the symmetric matrix
$\Ga^{1/2}\CF\Ga^{1/2}=\sum_{a=1}^T \delta_a \bm{v}_a\bm{v}_a\T$, one then
shows in a few lines that (\ref{eps_asympt1}) can be written as
\be
\epsilon_{\tau}\approx
\gamma_\tau^{-1} {\textstyle\sum_a} (v_{a,\tau})^2 \delta_a g(n\delta_a)
\label{asympt}
\ee
where $g(h)=\tr(\La^{-1}+h)^{-1} = \sum_i
(\lambda_i^{-1}+h)^{-1}$ and $v_{a,\tau}$ is the $\tau$-th component
of the $a$-th eigenvector $\bm{v}_a$. This is the general asymptotic form of our
prediction for the average Bayes error for task $\tau$.

To get a more explicit result, consider the case where sample
functions from the GP prior have (mean-square) derivatives up to order
$r$. The kernel eigenvalues $\lambda_i$ then decay as\footnote{ See
  the discussion of Sacks-Ylvisaker conditions in
  e.g.~\cite{WilRas06}; we consider one-dimensional inputs here though
  the discussion can be generalized.}  $i^{-(2r+2)}$ for large $i$,
and using arguments from~\cite{SolHal02} one deduces that $g(h)\sim
h^{-\alpha}$ for large $h$, with $\alpha=(2r+1)/(2r+2)$. In
(\ref{asympt}) we can then write, for large $n$, $g(n\delta_a)\approx 
(\delta_a/\gamma_\tau)^{-\alpha}g(n\gamma_\tau)$ and hence
\be
\epsilon_{\tau}\approx g(n\gamma_\tau)
\{{\textstyle\sum_a} (v_{a,\tau})^2 (\delta_a/\gamma_\tau)^{1-\alpha}
\}
\label{asympt2}
\ee
When there is only a single task, $\delta_1=\gamma_1$ and this
expression reduces to
$\epsilon_1=g(n\gamma_1)=g(n_1/\sigma_1^2)$. Thus
$g(n\gamma_\tau)=g(n_\tau/\sigma_\tau^2)$ is the error we would get by
ignoring all examples from tasks other than $\tau$, and the term in
$\{\ldots\}$ in (\ref{asympt2}) gives the ``multi-task gain'', i.e.\
the {\em factor} by which the error is reduced because of examples from
other tasks. 
(The {\em absolute}
error reduction always vanishes trivially for $n\to\infty$, along with
the errors themselves.)
One observation can be made directly. Learning of very
smooth functions, as defined e.g.\ by the SE kernel, corresponds to
$r\to\infty$ and hence $\alpha\to 1$, so the multi-task gain tends to unity:
multi-task learning is asymptotically useless. The only exception
occurs when some of the tasks are fully correlated, because one or
more of the eigenvalues $\delta_a$ of $\Ga^{1/2}\CF\Ga^{1/2}$ will
then be zero.

Fig.~\ref{fig:SE_useless}(left) shows this effect in action, plotting
Bayes error against $\rho^2$ for the two-task setting of
Fig.~\ref{fig:SE_Gaussian}(left) with $n=500$. Our predictions capture
the nonlinear dependence on $\rho^2$ quite well, though the effect is
somewhat weaker in the simulations. For larger $n$ the predictions
approach a curve that is constant for $\rho<1$, signifying negligible
improvement from multi-task learning except at $\rho=1$. It is worth
contrasting this with the lower bound from~\cite{Chai09}, which is
{\em linear} in $\rho^2$. While this provides a very good
approximation to the learning curves for moderate $n$~\cite{Chai09},
our results here show that asymptotically this bound can become very loose.

When predicting rough functions, there is some asymptotic improvement
to be had from multi-task learning, though again the multi-task gain
is nonlinear in $\rho^2$: see Fig.~\ref{fig:OU_useless}(left, inset)
for the OU case, which has $r=1$). A simple expression for the gain
can be obtained in the limit of many tasks, to which we turn next.

\subsection{Many tasks}

We assume
as for the two-task case that all inter-task correlations,
$\Cf_{\tau,\tau'}$ with $\tau\neq \tau'$, are equal to $\rho$, while
$\Cf_{\tau,\tau}=1$. This setup was used e.g.\ in~\cite{RodDen10}, and
can be interpreted as each task having a component proportional to
$\sqrt{\rho}$ of a shared latent function, with an independent
task-specific signal in addition. We assume for simplicity that we
have the same number $n_\tau=n/T$ of examples for each task, and that
all noise levels are the same, $\sigma_\tau^2= \sigma^2$. Then also
all Bayes errors $\epsilon_\tau=\epsilon$ will be the same. Carrying
out the matrix inverses in (\ref{eps_sol2}) explicitly, one can then
write this equation as
\be
\epsilon=g_T(n/(\sigma^2+\epsilon),\rho)
\label{eps_cond_largeT}
\ee
where
$g_T(h,\rho)$ is related to the single-task function $g(h)$ from above by
\be
g_T(h,\rho) = \frac{T-1}{T}(1-\rho)g(h(1-\rho)/T) +
\left(\rho+\frac{1-\rho}{T}\right) g(h[\rho+(1-\rho)/T])
\label{gT}
\ee
Now consider the limit $T\to\infty$ of many tasks. If $n$ and hence
$h=n/(\sigma^2+\epsilon)$ is kept fixed, $g_T(h,\rho)\to (1-\rho)+\rho
g(h\rho)$; here we have taken $g(0)=1$ which corresponds to
$\tr\La=\langle C(x,x)\rangle_x=1$ as in the examples above. One can
then deduce from (\ref{eps_cond_largeT}) that the Bayes error for any
task will have the form 
$\epsilon = (1-\rho)+\rho\tilde\epsilon$, where $\tilde\epsilon$
decays from one to zero with increasing $n$ as for a {\em single} task,
but with an effective noise level $\tilde
\sigma^2=(1-\rho+\sigma^2)/\rho$.  Remarkably, then, even though here
$n/T\to 0$ so that for most tasks no examples have been seen, the
Bayes error for each task decreases by ``collective learning'' to a
plateau of height $1-\rho$. The remaining decay of $\epsilon$ to zero
happens only once $n$ becomes of order $T$. Here one can show, by
taking $T\to\infty$ at fixed $h/T$ in (\ref{gT}) and inserting into (\ref{eps_cond_largeT}), that
$\epsilon=(1-\rho)\bar{\epsilon}$ where $\bar\epsilon$ again decays as
for a single task but with an effective number of examples $\bar{n}=n/T$ and
effective noise level $\bar\sigma^2/(1-\rho)$. This final stage of
learning therefore happens only when each task has seen a considerable
number of exampes $n/T$. Fig.~\ref{fig:many_tasks}(right) validates
these predictions against simulations, for a number of tasks ($T=200$)
that is in the same ballpark as in the many-tasks application example
of~\cite{Heskes98}. The inset for $T=1000$ shows clearly how the two
learning curve stages separate as $T$ becomes larger.

Finally we can come back to the multi-task gain in the asymptotic
stage of learning. For GP priors with sample functions with derivatives up to
order $r$ as before, the function $\bar\epsilon$ from above will decay
as $(\bar{n}/\bar{\sigma}^2)^{-\alpha}$; since
$\epsilon=(1-\rho)\bar\epsilon$ and
$\bar\sigma^2=\sigma^2/(1-\rho)$, the Bayes error $\epsilon$ is then
proportional to $(1-\rho)^{1-\alpha}$. This multi-task gain again
approaches unity for $\rho<1$ for smooth functions
($\alpha=(2r+1)/(2r+2)\to 1$). Interestingly, for rough functions
($\alpha<1$), the multi-task gain decreases for small $\rho^2$ as
$1-(1-\alpha)\sqrt{\rho^2}$ and so always lies {\em below} a linear
dependence on $\rho^2$ initially. This shows that a
linear-in-$\rho^2$ lower error bound cannot generally apply to
$T>2$ tasks, and indeed one can verify that the derivation
in~\cite{Chai09} does not extend to this case.

\section{Conclusion}

We have derived an approximate prediction (\ref{eps_sol2}) for
learning curves in multi-task GP regression, valid for arbitrary
inter-task correlation matrices $\CF$. This can be evaluated
explicitly knowing only the kernel eigenvalues, without sampling or
recourse to single-task learning curves. The approximation shows that
pure transfer learning has a simple lower error bound, and provides a
good qualitative account of numerically simulated learning
curves. Because it can be used to study the asymptotic behaviour for
large training sets, it allowed us to show that multi-task
learning can become asymptotically useless: when learning smooth
functions it reduces the asymptotic Bayes error only if tasks are fully
correlated. For the limit of many tasks we found that, remarkably, some
initial ``collective learning'' is possible even when most tasks have not seen
examples. A much slower second learning stage then requires many examples
per task. The asymptotic regime of this also showed explicitly that a lower
error bound that is linear in $\rho^2$, the square of the inter-task
correlation, is applicable only to the two-task setting $T=2$.

In future work it would be interesting to use our general result to
investigate in more detail the consequences of specific choices for the
inter-task correlations $\CF$, e.g.\ to represent a lower-dimensional
latent factor structure. One could also try to deploy similar
approximation methods to study the case of model mismatch, where
the inter-task correlations $\CF$ would have to be learned from data. More
challenging, but worthwhile, would be an extension to multi-task
covariance functions where task and input-space correlations to not
factorize.


\small{
\bibliographystyle{unsrt}
\bibliography{/home/psollich/references/references.bib}
}

\end{document}